\documentclass[letterpaper,10pt,conference]{ieeeconf}  % Comment this line out if you need a4paper

\IEEEoverridecommandlockouts                              % This command is only needed if 
\usepackage{times}
\usepackage{epsfig}
\usepackage{graphicx}
\usepackage{amsmath}
\usepackage{amssymb}
\usepackage{kotex}
\usepackage{kotex-logo}
\usepackage[linesnumbered,ruled]{algorithm2e}
\usepackage{caption}
\usepackage{subcaption}
\usepackage{multirow}
\usepackage{url}
\usepackage{caption}
\usepackage{subcaption}
\usepackage{url,gensymb}
\newsavebox{\measurebox}
\usepackage{caption}
\usepackage{subcaption}
\usepackage{amsmath}
\usepackage{multirow}

\title{\LARGE \bf A Single Multi-Task Deep Neural Network with Post-Processing for Object Detection with Reasoning and Robotic Grasp Detection}

\author{Dongwon Park$^1$, Yonghyeok Seo$^1$, Dongju Shin, Jaesik Choi and Se Young Chun*% <-this % stops a space	\thanks{}% <-this % stops a space
	\thanks{Dongwon Park, Yonghyeok Seo, Dongju Shin and Se Young Chun are with School of Electrical and Computer Engineering, UNIST, Ulsan, Republic of Korea. Jaesik Choi is with Graduate School of Artificial Intelligence, KAIST, Daejeon, Republic of Korea.}
	\thanks{$^1$Equal contribution. *Corresponding author: sychun@unist.ac.kr}%
}

\begin{document}
	
	\maketitle
	\thispagestyle{empty}
	\pagestyle{empty}

	%%%%%%%%%%%%%%%%%%%%%%%%%%%%%%%%%%%%%%%%%%%%%%%%%%%%%%%%%%%%%%%%%%%%%%%%%%%%%%%%
	\begin{abstract}
Recently, robotic grasp detection (GD) and object detection (OD) with reasoning have been investigated using deep neural networks (DNNs). There have been works to combine these multi-tasks using separate networks so that robots can deal with situations of grasping specific target objects in the cluttered, stacked, complex piles of novel objects from a single RGB-D camera.
We propose a single multi-task DNN that yields the information on GD, OD and relationship reasoning among objects with a simple post-processing.
%	Robotic grasping detection for novel objects is a challenging task. Grasping robotic single detection in virtual environment have achieved performance improvements, up to 97\% accuracy. But, enhancing performance of grasping real objects problems are still ongoing. In addition, how to let robots grasp specific objects in clutter and stacking scenes is still open problems. We propose highly real time, fully convolutional neural network(FCNN) based methods for grasping robotic single and multiple objects. 
Our proposed methods yielded state-of-the-art performance with the accuracy of 98.6\% and 74.2\% and the computation speed of 33 and 62 frame per second on VMRD and Cornell datasets, respectively. Our methods also yielded 95.3\% grasp success rate for single novel object grasping with a 4-axis robot arm and 86.7\% grasp success rate in cluttered novel objects with a Baxter robot.
%their performance in various situations with  and a Baxter robot.
%respectively) with state-of-the-art real-time computation time
%is well-integrated vision-based robot grasping detection and real robotic system, yielding state-of-the-art accuracy (up to 98.6\%, 72.28\%, respectively) with state-of-the-art real-time computation time for high-resolution images (33FPS, 62FPS) on VMRD and Cornell dataset, respectively.
	\end{abstract}

	%%%%%%%%%%%%%%%%%%%%%%%%%%%%%%%%%%%%%%%%%%%%%%%%%%%%%%%%%%%%%%%%%%%%%%%%%%%%%%%%
	\section{INTRODUCTION}
	
Robot grasping of particular target objects in cluttered, stacked and complex piles of novel objects is a challenging open problem. % in robotics.
Humans instantly identify / locate target objects and their nearby objects (object detection or OD), figure out location-wise relationship among objects (reasoning), and detect multiple grasps of the targets and their associated objects (grasp detection or GD).
%plan how to pick them up orderly (planning) and actually grasp them reliably (control). 
However, these tasks are still quite challenging for robots. 
Locating the targets and nearby objects in the piles of objects, reasoning their relationships %the picking orders of them 
and detecting multiple robotic grasps accurately and quickly are important multi-tasks for successful robotic grasping.
%	Intelligent robotic grasping detection은 매우 어려운 task 입니다. 일상속에서 intelligent robotic grasping detection을 사용하기 위해서는 3가지 조건을 만족해야 합니다. 
%For this goal, it is required to accurately detect grasps, to comprehensively reason the relationships among the target and its related objects in cluttered or stacked objects, and to compute all of them in real-time.
%	첫번째, grasp detection이 정확해야 합니다. 두번째, 복잡한 상황(cluttered object or stacking object)을 인식하여, 우리가 원하는 target만을 grasping 해야 합니다. 세번째, 실시간으로 구동이 가능해야 합니다. 
%In this way, there are multi-task problems to use the robot in daily life.
%	Humans instantly identify multi-task grasping areas with object detection and relationship inter-objects of clutter objects or stacking object (perception and reasoning), almost instantly plan how to pick them up(planning) and then actually grasp it reliably (control). However, accurate Multi-task robotic grasp detection, trajectory planning, and reliable execution are quite challenging for robots. 

	\begin{figure}[!b]
	\vskip -0.15in
	\centering
	\includegraphics[width=1.0\linewidth]{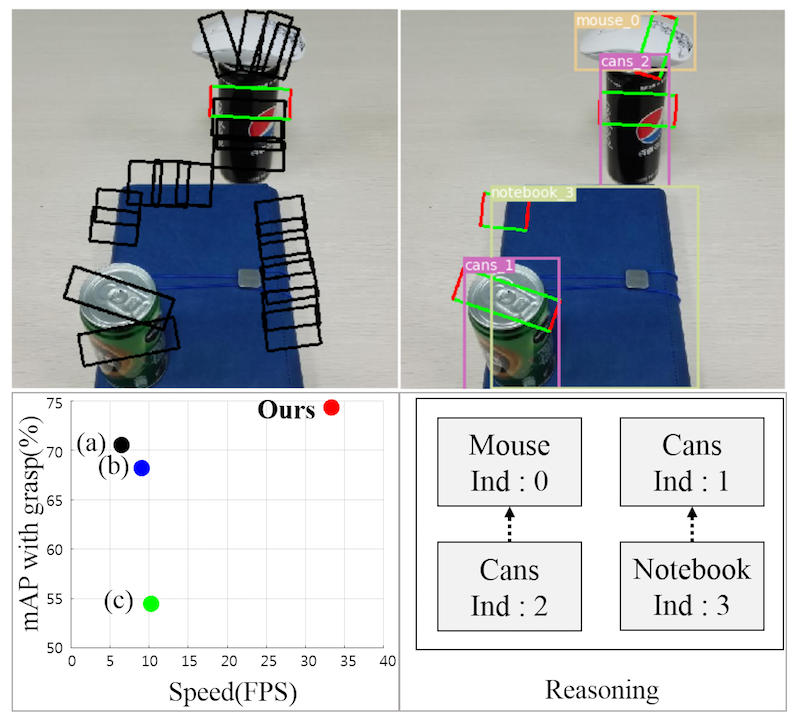}
	\caption{(top left panel) GD with grasp candidates (black rectangles) and the best grasp (green and red rectangle)
%	(only using the grasp detection), green and red means best grasp and black means others. 
(right panels) multi-tasks of GD, OD and relationship reasoning. (bottom left panel)
computation speed (FPS) vs. prediction accuracy (mAP) for multi-task grasping detection of
our method achieving state-of-the-art performance in both accuracy and speed
and other previous works of (a)~\cite{zhang2018rprg}, (b,c)~\cite{zhang2018roi}.
%Our proposed methods achieve both state-of-the-art prediction accuracy and the fastest computation time.
}
	\label{fig:intro}
\end{figure}

Deep learning based approaches have been actively investigated for robot grasp detection since the work of Lenz \textit{et al.}~\cite{Lenz:2013uz,Lenz:2015ih}. Thanks to the Cornell robotic grasp detection open database~\cite{Lenz:2013uz} and the advance of deep learning techniques, there have been many works proposed~\cite{Redmon:2015eq,guo2017hybrid,Chu:ek} and current state-of-the-art GD accuracy on the Cornell dataset is up to 97.7\%~\cite{zhou2018fully}. The Cornell dataset contains images with a single object and multiple grasp labels. Deep neural networks (DNNs) trained with this data generally yielded multiple grasps for a single object or multiple objects that are separately placed. Due to the capacity of DNNs, they often yielded good GD results for novel objects. 
However, these DNNs with the Cornell dataset do not seem general enough for the task of robotic grasping in the cluttered piles of objects.
%	딥러닝의 발달로 robot grasp detection에서 많은 발전이 있었습니다.~\cite{Lenz:2013uz,Lenz:2015ih,Redmon:2015eq,guo2017hybrid,Chu:ek,zhou2018fully} 최근, robot grasping data with single object인 Cornell에서 정확도는 97.7\%를 기록하였습니다~\cite{zhou2018fully}. 
%	그러나, robot grasping point만 예측을 하기 때문에 Multi object problem에서 한계가 존재합니다.
Moreover, it is especially challenging which object the robot has to grasp first in order not to damage other objects for cluttered or stacked objects or
in order to efficiently grasp specific target objects.
In other words, a robot must know if an object is on another object in the piles of objects for successful grasping.

%	robot grasping with single object 모델을 이용하여 Multi-object를 예측하게 되면, 단순히 해당 입력에서의 grasping point만를 예측하기 떄문에 clutter object or stacking object에서 target object를 grasp하는것은 한계가 있습니다.
	
Recently, Zhang \textit{et al.} proposed multi-task convolution robotic grasping networks to address the problem of combining GD and OD with relationship reasoning in the piles of objects~\cite{zhang2018rprg}. This method consists of several DNNs that are responsible for
generating local feature maps, GD, OD and relationship reasoning separately.
%	이러한 문제를 해결하기 위하여, Zhang은 Multi-task Convolution robotic grasping network를 제안 하였습니다.
%    Multi-task robotic grasping consists of perception, reasoning and grasping. 
 %   In perception, both an object and robotic grasps on that object are estimated.
  %  In reasoning, the relationships among objects are predicted to yield grasping order. 
   % Lastly, in grasping, the predicted relationships among objects are considered to perform graspings.
	%Multi-task robotic grasping는 Perception, Reasoning and Grasp으로 구성되어 있습니다. In perception part, object와 object에 대한 grasping point를 예측 합니다. In reasoning part, 물체들간의 relationships을 예측하며, 예측된 relationships를 이용하여, grasping order를 얻을 수 있습니다. In grasp part, Reasoning에서 얻은 물체간의 relationship을 반영하여, 실제 grasping을 합니다. it is an important task for successful multi-task robotic grasping. ~\cite{zhang2018rprg}
%    Each step is an important task for successful multi-task robotic grasping.
    More specifically, features are extracted using ResNet-101 based region proposal network (RPN) and then are fed into three DNNs corresponding to three tasks: OD, GD and relationship prediction among objects to perform grasping considering relationships and orderings (reasoning).
%	그들은 Resnet-101기반인 RPN을 이용하여, Feature를 추출 한후 3가지 networks~\cite{zhang2018rprg}를 통해, Object detection , grasp detection, object간에 relationship을 예측하여, relationship이(reasoning) 반영된 grasping을 하였습니다. 
	This approach allowed easy matching for reasoning and achieved 
	high GD accuracy of 70.1\% on the VMRD robot grasping dataset~\cite{zhang2018visual}
	with reasonable computation speed of 6.5 frame per second (FPS). 
	However, %it seems desirable to achieve higher accuracy and faster computation speed.
	this modular structure could be further optimized and improved for higher accuracy, faster computation speed and less DNNs for potentially reduced GPU memory usage.
%	since large feature maps are extracted and separate detection networks are used, computation speed seems relatively slow (6.5 frame per second or FPS).
%	RPN을 이용하여 각 물체에 대한 feautre를 추출 한 후, object, grasp, relationship을 predict 하기 때문에, 쉽게 매칭시킬수 있으며, 정확도가 높은 장점이 있습니다. 그러나, 많은 feature를 추출 후, Detection networks를 사용하기 때문에, 처리시간이 오래걸리는 단점이 존재합니다.
	%There are still remaining challenges in robot grasping even though current state-of-the-art object detection with grasp detection accuracy of 70.1\% with reasonable computation speed of 6.5 FPS seems impressive~\cite{zhang2018rprg}.
  %  Achieving faster computation time for generating multi-task grasps with state-of-the-art accuracy seems desirable especially for multi-task robotic grasping of real-environment. 
  %  Thus, it seems desirable to reduce computation time while maintaining (or improving) multi-task grasp detection accuracy.
    
	In this paper, we propose a single multi-task DNN with a simple post-processing for OD with reasoning and GD on the piles of novel objects using the information from a single RGB-D camera.
	%highly real-time, novel instance one-stage convolutional neural network based methods for multi-task robotic grasp detection from RGB images based on a state-of-the-art object detection network, YOLOv3~\cite{redmon2018yolov3}, 
	Our proposed methods extended YOLOv3~\cite{redmon2018yolov3}, a state-of-the-art OD method, to dealing with multi-tasks of OD, GD and relationship reasoning, but maintained its simple single network structure.
	Ablation studies were performed to further optimize %better understand the properties of 
	different components %(predictions across scale, class activation, optimizer, loss function for object class) 
	of our multi-task networks.
%	by incorporating several important components for multi-task robot grasping detection on stacking objects.
Our methods yielded state-of-the-art multi-task GD performance (74.2\%, 98.6\%) on the VMRD and Cornell datasets, respectively, with real-time computation speed (30 and 62 FPS) for high-resolution images of $608 \times 608$ and $320 \times 320$ as illustrated in Fig.~\ref{fig:intro}.
%grasp detection and grasp detection performance on the VMRD dataset and Cornell dataset(Image-wise) in almost all performance metrics with highly real-time computation speed(30 FPS and 62 FPS) even for high-resolution images ($608 \times 608$ and $320 \times 320$ image) as illustrated in Fig.~\ref{fig:intro}.
%We also carefully investigated the properties of our proposed methods by ablation studies to better understand the properties of different components(across scale, class activation, optimizer, loss function for object class) for multi-task robot grasp detection algorithms.
Our methods were also applied to real robotic grasping tasks with a 4 axis robot arm on single novel objects as well as 
a Baxter on multiple novel objects in various settings of piles (cluttered, stacking, invisible scenes)
and
yielded 95.3\% grasp success rate for single novel object grasping and 86.7\% grasp success rate in cluttered novel objects, respectively.

\section{RELATED WORK}

\noindent \textbf{Pre-deep learning era.}
Data-driven GD for novel objects has been investigated extensively~\cite{Bohg:2014ef}.
Saxena \textit{et al.} proposed a machine learning (ML) based method to rank 
the best graspable location for all candidate image patches from different locations~\cite{Saxena:2008doa}.
Jiang \textit{et al.} proposed a 5D robotic grasp representation by using %proposing 
a ML method to rank the best
graspable image patch whose representation includes orientation and gripper distance 
among all candidates~\cite{Jiang:2011ja}.

\noindent \textbf{Two-stage classification based approach.}
Lenz \textit{et al.} proposed to use a sparse auto-encoder (SAE) %, an early deep learning model, %
to train the network 
to rank the best graspable candidate image patch from sliding window
with RGB-D~\cite{Lenz:2013uz,Lenz:2015ih}.
Wang \textit{et al.} proposed a real-time classification based GD method 
using a stacked SAE for classification with efficient grasp candidates generation~\cite{Wang:2016cpa}.
%It also reduced the number of grasp representation parameters %to estimate 
%such as height for known gripper and orientation.
%that could be analytically obtained from surface norm.
Mahler \textit{et al.} proposed Dex-Net 2.0 with GQ-CNN that estimates grasps for parallel grippers from a single depth image trained with synthetic point-cloud data~\cite{Mahler:2017te}.

\noindent \textbf{Single-stage regression based approach.} 
	Redmon \textit{et al.} proposed a deep learning regressor GD method based on the AlexNet~\cite{Krizhevsky:2012wl} %that 
%	that yielded 84.4\% (image-wise) and 84.9\% (object-wise) 
	with fast computation time~\cite{Redmon:2015eq}.
	When performing
	robotic grasp regression and object classification together, image-wise prediction accuracy
%	of 85.5\% 
    was able to be improved without increasing computation time.
	Kumra \textit{et al.} also proposed % and Kanan proposed 
	a real-time regression based GD method using
	ResNet~\cite{He:2016ib} especially for RGB-D and this work yielded improved performance with fast computation time.
%	Their method yielded up to 89.2\% (image-wise) and 88.9\% (object-wise) prediction accuracies
%	with fast computation time (103 ms per image)~\cite{Kumra:2017ko}.

\noindent \textbf{Multibox based approach.} Redmon \textit{et al.} also proposed a multibox  GD method (called MultiGrasp) by dividing the input image into S$\times$S grid and applying 
	regression based GD to each grid box~\cite{Redmon:2015eq}. 
	This approach did not increase computation time %(76 ms per image), 
	but increased prediction accuracy for multiobject, multigrasp detection. % up to 88.0\% (image-wise) and 
	Guo \textit{et al.} proposed a hybrid multibox approach with visual and tactile information %based on ZF-net~\cite{Zeiler:2014fr}
	by classifying graspability, angles ($\theta$), and by regressing locations and grasp width ($w$), height ($h$)~\cite{guo2017hybrid}. 
	Chu \textit{et al.} proposed two-stage neural networks combining
	grasp region proposal network (RPN) and robotic GD network~\cite{Chu:ek}.
	Zhou \textit{et al.} proposed rotational anchor box and angle matching (angle classification + regression) to further improve prediction accuracy~\cite{zhou2018fully}. This method is currently state-of-the-art in GD on the Cornell dataset % with deep ResNet-100~\cite{He:2016ib} and 
	with computation time of 117ms per image (320$\times$320). 
	Note that the networks of Guo~\cite{guo2017hybrid}, Chu~\cite{Chu:ek} and Zhou~\cite{zhou2018fully} are based on faster-RCNN~\cite{Ren:2015ug}. %, a two-stage network.
    
\noindent  \textbf{Hybrid approach.}
	Asif \textit{et al.} proposed GraspNet that predicts graspability and then estimates robotic grasp
	parameters based on high-resolution grasp probability map~\cite{Asif:2018ud}.

\noindent  \textbf{Depth vs color information.}
 There are several works that use depth information only or color information only for GD. Johns \textit{et al.} developed a method to estimate a grasp score (quality) from a single depth image~\cite{Johns:2016gm}.
Dex-Net 3.0 was proposed to estimate robotic grasps for suctions from a depth image (point cloud) trained with synthetic data~\cite{Mahler:2018kn}.
There have been a couple of works to use depth images only for closed-loop grasping~\cite{Viereck:2017uq,morrison2018closing}.
Morrison \textit{et al.} demonstrated that using fast, lightweight neural network was important for grasping dynamic objects~\cite{morrison2018closing}.
There also have been some works using color images only for GD. Since depth image is often quite noisy~\cite{Khoshelham:2012jl},
only RGB images have been used for learning 5D grasp representation from a color image~\cite{Pinto:2016iu} and for achieving almost state-of-the-art performance~\cite{zhou2018fully}.

\noindent \textbf{OD with GD.}
Zhang \textit{et al.} proposed a VMRD grasping dataset with object detection and object relation and a Visual manipulation relationship network (VMRN)~\cite{zhang2018visual}. Based on SSD~\cite{liu2016ssd}, an OD method, VMRN extracted features and then predicted relationship of objects.
Zhang \textit{et al.} further developed multi-task robotic grasp networks for OD, GD and reasoning with VMRN~\cite{zhang2018roi,zhang2018rprg} based on the GD work of Zhou~\cite{zhou2018fully} for grasping tasks in complex piles of objects.
 %Zhang ~\cite{zhang2018rprg} based on their previous model~\cite {zhang2018roi}, proposed a multi-task robotic grasp network using Manipulation relationship predictor that predicts the relationship between objects. Additionally Zhang~\cite{zhang2018rprg} also proposed a network which predict an object detection and graspipng detection together in reference to Zhou~\cite{zhou2018fully} by adding an RPN module. Therefore, Zhang ~\cite{zhang2018rprg} can conduct some tasks in object complex scene.

\section{PROPOSED METHODS}

\subsection{Problem description}

\noindent \textbf{Single object robot grasping} A 5D robotic grasp representation
is widely used for GD with a parallel gripper
when a single 2D image (RGB or RGB-D) is used~\cite{Jiang:2011ja,Lenz:2015ih}.
This representation is a vector of 
$\{x_{gd}, y_{gd}, \theta_{gd}, w_{gd}, h_{gd}\}$ that consists of
%The goal of the problem is to predict 5D robotic grasp representations~\cite{Jiang:2011ja,Lenz:2015ih}
%for single object from a given color image (RGB) and possibly depth image (RGB-D)
%where a 5D robotic grasp representation consists of 
location $(x_{gd}, y_{gd})$,
orientation $\theta_{gd}$,
gripper opening width $w_{gd}$ and 
parallel gripper plate size $h_{gd}$.
%Then, the 5D robotic grasp representation 
%in camera vision coordinate system 
%should be transformed into a new 5D grasp representation 
%\(
%\{ \tilde{x_{gd}}, \tilde{y_{gd}}, \tilde{\theta_{gd}}, \tilde{w_{gd}}, \tilde{h_{gd}} \}
%\)
%in robot coordinate system so that it can be used for actual robot grasping task.
 
\noindent \textbf{Multi-task robot grasping}

Grasping a specific target object in cluttered and stacking objects requires more than single object grasping information and needs additional information such as object class and relationship reasoning (see Fig.~\ref{fig:intro}) for sequential grasp planning.
%The goal of multi-task problem is to perform predict the best way robotic grasping detection for a specific target in cluttered and stacking objects. To grasp a exact target object in those scene, robots predict not only 5D robotic grasp representation but also the class and grasping order for each objects.
%Grasping order means order of object stacking level by recognizing stacking relation inter-objects through a graph as illustrated in Fig~\ref{fig:intro}(Right). 
We extended the previous 5D robotic grasp representation to including object class and stacking order among objects as follows:
%Classes stand for object class labels about a predicted optimal grasping point.
%Utilizing above 7D robotic grasp representations, robot prioritize stacking objects. After distinguishing between other objects from a target, robots move the other objects and are able to grasp the target more accurately.
% 물체가 겹겹히 쌓여있는 환경(cluttered object) 혹은 이런 환경에서 원하는 물체를 파지하고자 하는 것이 Multi-task robot grasping입니다.
% Stacking된 물체들에 대해 grasping을 하기 위해선, 각 물체에 대한 5D robotic grasp representation 이외에 추가적으로 (grasping order, class)를 예측해야 합니다.
%grasping order는 objects간에 stacking 관계를 graph를 통하여 인지하여, 우선적으로 grasping 할 물체의 순서를 의미합니다. class는 예측 된 grasping point에 대한 해당 물체의 class를 의미 합니다.
%이것을 통하여, 우리는 stacking된 물체에 대해 우선순위를 주어, 순차적으로 물체를 grasp 할 수 있고, 이에 더하여 우리가 grasp하기 원하는 물체와 장애물을 인식하여 장애물을 치우고, 보다 안전하게 grasp을 할 수 있습니다.
%Then, The 7D robotic grasp representation as follows:
\[
\{ x_{gd}, y_{gd}, \theta_{gd}, w_{gd}, h_{gd}, cls_{gd}, ord_{gd} \}.
\]

\subsection{Reparametrization of 15D representation} 

We propose a 15D representation for multi-task robot grasping problem to exploit a single multi-task DNN for OD, GD and reasoning altogether. 
The parameters for OD are %Predicting parameters of solving OD problem are 
$\{x_{od}, y_{od}, w_{od}, h_{od}, cls_{od}, pr_{od}\}$ and the parameters for GD are
%The parameters of GD are 
$\{x_{gd}, y_{gd}, w_{gd}, h_{gd}, cls_{gd}, pr_{gd}, \theta_{gd}\}$. 
where $pr_{od}$ is a probability of an object existing and $pr_{gd}$ is a graspable probability.
The parameters of reasoning are $\{cls_{fc}, cls_{cc}\}$ for ordering objects ($ord_{gd}$).
%for the $ord_{gd}$. The 
Father class (FC) and children class (CC) are labels under and over the predicted target object, respectively. FC and CC are predicted of each grid.
%우리는 이러한 문제를 해결 하기 위해 Object Detection(OD), Grasp Detection(GD), Reasinging으로 문제를 세분화 하였습니다. 
%Object detection문제해결을 위해 예측하는 파라미터는  $\{x^{od}, y^{od}, w^{od}, h^{od}, class^{od}, probability^{od}\}$ 입니다. Grasping detection에서 예측하는 파라미터는 $\{x^{gd}, y^{gd}, w^{gd}, h^{gd}, class^{gd}, probability^{gd}, \theta^{gd}\}$ 입니다. Reasoning에서 우리가 예측하는 파라미터는 $\{father classes, children classes\}$ 입니다. father class란 object detection에서 예측된 물체, 그 '아래'에 있는 다른 물체의 class를 의미 합니다. children class란, object detection에서 예측된 물체 '위에' 있는 물체의 object class를 의미합니다. father class와 children class는 objcet detection에서 각 grid 마다 예측하도록 하였습니다.

We propose the following reparametrization of OD and GD for robotic grasping in the piles of objects:
\begin{eqnarray*}
	OD = \{t^x_{od}, t^y_{od}, t^w_{od}, t^h_{od}, t^{pr}_{od}, t^{cls}_{od}\}, 
	R = \{t^{cls}_{fa}, t^{cls}_{cc}\} \\
	GD = \{t^x_{gd}, t^y_{gd}, t^w_{gd}, t^h_{gd}, t^{pr}_{gd}, t^{cls}_{gd}, t^{\theta}_{gd}\}
\end{eqnarray*}
% z in upper sign in od gd edit syh // there is no picture in fig 2 for z
where 
\(
x_{j} = \sigma( t^x_{j} ) + c^x_{j}, y_{j} = \sigma( t^y_{j} ) + c^y_{j},
\)
$\sigma(\cdot)$ is a sigmoid function, 
\(
w_{j} = p^{w}_{j} \exp( t^w_{j} ), h_{j} = p^h_{j} \exp( t^h_{j} ), \theta_{gd} = p^{\theta}_{gd} +  t^{\theta}_{gd}, 
cls_{j} = \mathrm{softmax}(t^{cls}_{j}), 
cls_{rs} = \sigma( t^{cls}_{rs} ), 
pr_{j} = \sigma( t^{pr}_{j} )
\), 
$j \in \{od, gd\}$ and $rs \in \{fc, cc\}$.
Note that %$\sigma( \cdot )$ is a sigmoid function, $\exp( \cdot )$ is an exponential function,
$p^h_{j}$, $p^w_{j}$ and $p^{\theta}_{gd}$ are the pre-defined height ,width, orientation of an anchor box, respectively, and 
$(c^x_{j}, c^y_{j})$ are the location of the top left corner of each grid cell (known).
Thus, DNN for GD of our proposed methods
will estimate $\{ t^x_{j}, t^y_{j}, t^{\theta}_{j}, t^w_{j}, t^h_{j}, t^{pr}_{j} \}$ instead of 
$\{ x_{j}, y_{j}, \theta_{gd}, w_{j}, h_{j}, {pr}_{j} \}$.
$x_{j}, y_{j}, w_{j}, h_{j}$ are properly normalized so that the size of each grid is $1 \times 1$.
Lastly, the angle $\theta_{gd}$ will be modeled as a discrete and continuous value instead of a continuous value. % that was also used in the work of Zhang~\cite{guo2017hybrid}.

\noindent \textbf{Anchor box: w, h in each cell.}
Anchor box approach has been used for OD~\cite{Redmon:2017gn}. 
Due to re-parametrization with anchor box, estimating $w_{j}, h_{j}$ is converted into estimating $t^w_{j}, t^h_{j}$, which are related to the expected values of various sizes of $w_{j}, h_{j}$. Then, %and then classifying 
the best grasp representation among all anchor box candidates is selected for the final output. Thus, %In other words, this 
re-parametrization changes regression problem into regression + classification problem for $w_{j}, h_{j}$. 

\noindent \textbf{Anchor box: orientation in each cell.}
While MultiGrasp took regression approach for $\theta_{gd}$~\cite{Redmon:2015eq}, Guo \textit{et al.} converted regression problem of estimating $\theta_{gd}$ into the classification for $\theta_{gd}$ among finite number of
angle candidates in $\{ 0, \pi/18, \ldots, 17\pi/18\}$~\cite{guo2017hybrid}. 
%Specifically, they predicted the $\theta_{gd} \in \{ 0, \pi/18, \ldots, 17\pi/18\}$. 
Zhang~\cite{zhang2017stereo} proposed orientation anchor box so that
the angle is determinded using classification as well as discrete anchor box rotations. %. Using this method, the angle was solved with Classification + Discrete. 
Mean average precision (mAP) increased by 3\% when using orientation anchor box (4 angles) over angle classification on the VMRD. % by using four orientation anchor boxes.
%By using Orientation Anchor box of the GD, we get more than 3\% improvement in mAP of the object detection with robot grasping .

%우리는 VMRD에서 Orientation Classification보다 Orientation Anchorbox를 사용하였을때, 3\% 정확도가 향상된다는것을 발견하였습니다.
%, Orientation Anchorbox를 사용 하였습니다. 4개의 Orientation anchorbox를 사용하였습니다.
%By using Orientation Anchor all of the convolutional layers in YOLO
%we get more than 2% improvement in mAP

\noindent \textbf{Object class: cls in each cell.} 
When objects are stacked in a complex way, it becomes a difficult task to match OD result (detection bounding box) with GD result without additional information such as object classes. For this task, object class is predicted for each of grasp detection box result so that 
%This is because the object detection box and grasping box lack a lot of information shared with each other. 
%To match robot grasp detection box with object detection box, the class is predicted from each of grasp detection box. Thus, 
our proposed model can yield %let 
grasping detection boxes, %know both 
their grasping points and corresponding object classes.
%Object detection box와 grasp detection box를 매칭하는것은 굉장히 어려운 일입니다. 그렇기 때문에 우리는 Robot grasp detection box와  object detection box를 매칭 시키기 위해서, 각 grasp detection box에서 class를 예측하도록 하여, 어떤 물체에 대한 grasping point인지 알 수 있도록 하였습니다. 또한, 하나의 이미지에서 같은 class의 object와 혼동하는것을 방지 하기 위하여, grasp detection box와 object detection box에 대한 IOU를 이용하여, 전처리를 하였습니다.
A softmax was selected for class activation function through our self-evaluation ablation study that will be reported shortly.
% Through self-evaluation study, we used softmax method on the class activation function. 
%여기서 기존의 YOLO-v3의 경우 Class activation function을 Sigmoid를 사용하였지만, Self Evaluation study를 통하여 Sigmoid 보다 Softmax가 적합하다는것을 알게 되었습니다. 

\noindent \textbf{FC and CC in each cell.} For inter-object relationship, we propose to predict %robot predict 
FC and CC along with other detection results. %with proposed model. 
FC and CC are class labels under and over the target object, respectively. 
FC and CC consist of object class labels and no-class label.
%fc와 cc는 기존 class와 다르게 class가 존재하지 않는다는 non class label를 기존 클래스에 추가적으로 사용하였습니다.
%우리는 물체간의 Relation을 알기 위해서, Father class와 Children class를 예측하도록 하였습니다. Father class(fc)란 내가 지정한 object가 어떤 물체 위에 있는지 말해주는 class를 의미 합니다. Children class(cc) 란 내가 지정한 object가 어떤 물체 아래에 있는지 말해주는 class를 의미 합니다. fc와 cc는 object class에서 non class label를 추가적으로 사용하였습니다.
$(t_{fc}^{cls}, t_{cc}^{cls}) \in \{class_1, class_2, ... , class_{none}\} $. 
In our experiment, we observed that CC is more accurately estimated than FC. 
Thus, we only used CC for reasoning for the best possible results. %the Children class.

\begin{figure}[!t]
	\centering
	\includegraphics[width=0.95\linewidth]{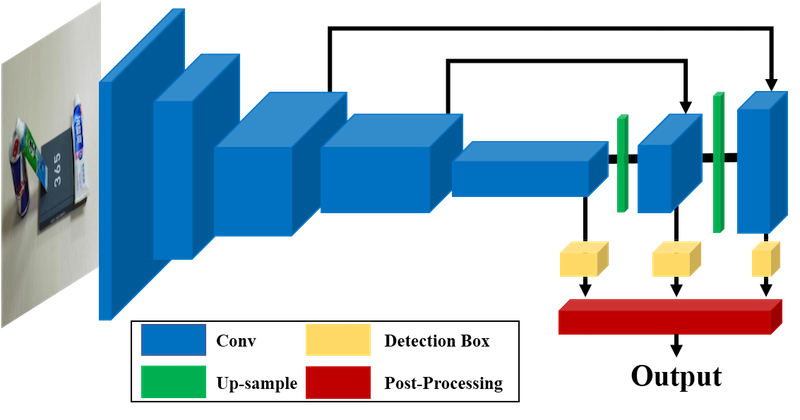}
	%		\vskip -0.1in
	\caption{Proposed FCNN architecture 
		based on Darknet.}
	\label{fig:yolo9000net}
	\vskip -0.15in
\end{figure}

\subsection{Proposed FCNN with predictions across scales} 

Our proposed FCNN inherited pre-trained Darknet-53 of YOLOv3 for OD~\cite{redmon2018yolov3} and extended it for multi-task OD with reasoning and GD as illustrated in Fig.~\ref{fig:yolo9000net}.
%We chose Darknet-19 as the base network for single object robot grasp detection and
%Darknet-53 as the base network for multi-task grasp detection as illustrated in Fig.~\ref{fig:yolo9000net}.  
%illustrates our proposed FCNN architecture based on Darknet-19 or Darknet-50. 
%Pre-trained networks were used for fast training and good performance. 
%Typical fully connected (FC) layers for image classification are replaced with $1 \times 1$ convolution layers were added to make FCNN architecture so that input images with any size ($e.g.$, high resolution images) can be processed. \\
For our multi-task predictions, we did not only adopted %used 
prediction across scales for OD using feature pyramid networks~\cite{lin2017feature}, but also extended it for reasoning and GD.
% that was used for object detection~\cite{redmon2018yolov3}. 

On the low-resolution scale, three anchor boxes $(w, h)$ for OD 1 anchor box for GD and 4 anchor boxes for grasping angles are predicted as
\begin{eqnarray*}
&& (p^{w}_{od}, p^{h}_{od}) \in \{ (540, 540), (480, 480), (420, 420) \}, \\
&& (p^{w}_{gd}, p^{h}_{gd}) \in \{ (300, 300) \},
p^{\theta}_{gd} \in \{ 0, \pi/4, 2\pi/4, 3\pi/4 \}.
\end{eqnarray*}
Then, on the mid-resolution scale after $\times$2 bilinear up-sampling, 3 anchor boxes for OD, 1 anchor box for GD and 4 anchor boxes for grasping angles are estimated as
%On the mid-level scale, the following anchor boxes are used:
\begin{eqnarray*}
&& (p^{w}_{od}, p^{h}_{od}) \in \{ (360, 360), (300, 300), (240, 240) \}, \\
&& (p^{w}_{gd}, p^{h}_{gd}) \in \{ (100, 100) \},
p^{\theta}_{gd} \in \{ 0, \pi/4, 2\pi/4, 3\pi/4 \}.
\end{eqnarray*}
On the high-resolution scale after $\times$4 up-sampling, the following anchor boxes for 
our multi-tasks are predicted:
%In object detection, for each scale, three anchor box(w, h) is predicted, one anchor box (w, h) predicted in grasp detection and four anchor box in orientation.
%There are two up-sampling blocks, so thus there are prediction boxes on three different scales.
%On the lowest scale, three anchor boxes are used as follows:
\[
(p^{w}_{od}, p^{h}_{od}) \in \{ (180, 180), (120, 120), (60, 60) \}.
\]
%YOLOv3 is using anchor boxes that are similar to feature pyramid networks~\cite{lin2017feature}. While YOLOv3 has three prediction boxes for $\times$1, $\times$2 and $\times$4 bilinear up-sampling on the object detection.

OD with reasoning are performed across scales of $\times$1, $\times$2 and $\times$4 and
GD are performed across scales of $\times$1 and $\times$2.
%We also used the same method for object detection with reasoning. But, we only use two $\times$1, $\times$2 Bilinear up-sampling and two grasp detection output for robot grasping. 
%각 scale 마다 object detehction에서 예측하는 Anchorbox(w,h)수는 3개 이며, grasp detection에서 예측하는 Anchorbox(w,h)수는 1개 이며, Anchorbox(orientation)수는 4개 입니다. 
Therefore, 9 anchor boxes are predicted with 4 bounding box offsets, object probability, object class (class number) and (class number+1)$\times$2 reasoning classes (FC, CC) for OD with reasoning. In addition, 8 anchor boxes are predicted with 4 bounding box offsets, orientation, grasp probability and object class (class number) for GD.

\subsection{Reasoning post-processing: from class to index}

\begin{figure}[!t]
        \begin{center}
		\centering
        \includegraphics[width=0.4\textwidth]{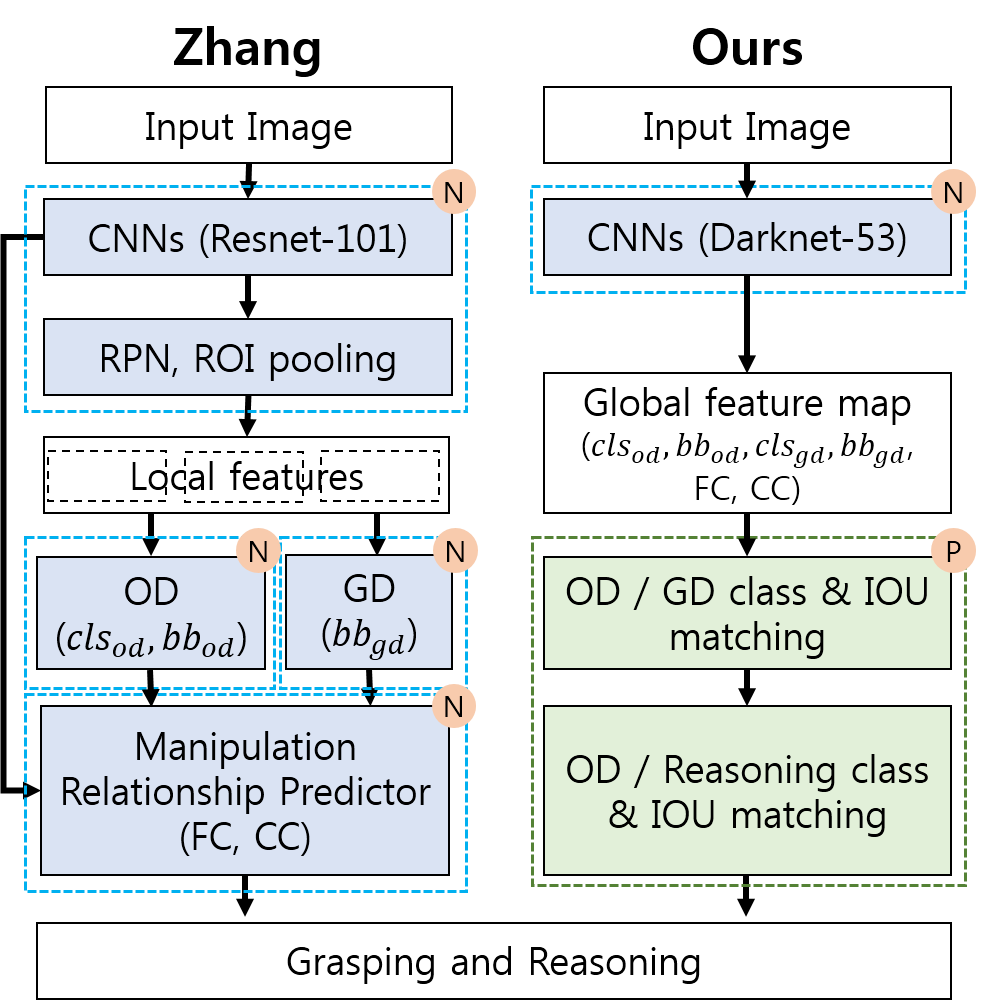}
        \caption{Schematic pipelines of Zhang~\cite{zhang2018rprg,zhang2018roi} vs ours.
        }
        \label{fig:compare}
        \end{center}
			\vskip -0.15in
\end{figure}
Fig.~\ref{fig:compare} illustrates the differences between the works of Zhang~\cite{zhang2018rprg,zhang2018roi} and our proposed methods. 
Previous works generated necessary information for OD with reasoning and GD. Deep neural networks (N) generated local features or OD or GD or relationship among objects (FC, CC), respectively and sequentially. However, as shown in the work of Redmon~\cite{Redmon:2015eq}, dealing with GD and classification often improves the overall performance of GD. We propose a novel single network (N) to yield most information on OD and GD with simple reasoning post-processing (P) for building hierarchy among objects.
%, it is easy to match OD and GD because OD and GD are performed after extracting the feature using RPN. However, our method does not use RPN, so post-processing is required in addition to Non-maximum suppression(NMS). 

For a generated global feature map including class information, bounding box information and FC / CC, reasoning post-processing can build index relationships using class information. Firstly, non-maximum suppression is applied to GD and OD to eliminate unnecessary detection results. Secondly, grouping of bounding boxes in OD and GD ($bb_{od}$, $bb_{gd}$) is performed based on their class information ($cls_{od}$, $cls_{gd}$). Then, the spatial information of bounding boxes are used for further grouping bounding box pairs for OD and GD based on the IOU (Intersection Over Union) as follows:
%So, we can build an index relationship by distinguishing objects of the same class based on the class information given in the network at this stage.
%Processing consists of NMS, combine OD with GD and reasoning. As mentioned earlier, we applied non-maximum suppression, which is frequently used in detection algorithm, to GD and OD to eliminate unnecessary detections. 
%Then do two things for OD and GD. First, match the object detection box and the grasping detection box by referring to their respective classes. 
%Second, now in matched candidates with class, to utilizing the spatial information of each boxes, we calculate the IOU for grasp detection box and object detection box. 
\[
IOU = \cfrac{bb_{od}\cap bb_{gd}}{bb_{od} \cup bb_{gd}}.
\]
Lastly, among GD candidates whose IOU exceeds a certain threshold, the best probability for GD is selected to obtain the final OD / GD bounding box pair.
%Then, among grasp detection candidates whose IOU exceeds a threshold, the best probability grasp detection is finally matched to the object detection box.
%In this way, the object detection accuracy is improved by 0.5\% (map). %In this work we got to know that those two detection model outputs can be improved complementary to each other.

Similarly, we compare object classes with child classes already obtained in the model to get the relationships between them by matching boxes with IOU threshold. With this, we can make a object relation graph for robot grasping.

\subsection{Loss function for multi-task GD, OD, reasoning} 

% that we will describe in the next subsection.
For the output vectors OD, GD and R of DNN and 
the ground truth (GT) $OD_{gt}$, $GD_{gt}$ and $R_{gt} $,
we propose the follow loss function to train our single multi-task DNN:
\begin{eqnarray*}
%	&& L(OD, GD, Reasoning) = \\
	&& \sum_{i\in \Omega} \sum_{j\in \{od, gd\}} \{ \sum_{k\in \{x, y, w, h\}} \mathrm{MSE}(k_j^{i},k_{j,gt}^{i}) + \\
    && \sum_{k\in {pr}} (-\log k_j^{i}) + \sum_{k\in {cls_{ob}}} \mathrm{FocLoss}(k_j^{i},k_{j,gt}^{i}) \} + \\
	&& \lambda_\mathrm{n} \sum_{i\in \Omega^c} \sum_{j\in \{od, gd\}} \sum_{k \in {pr}} (-\log (1-k_j^{i}))  + \\
%	&&  \sum^{P}_{i\in Pos} \sum^{K}_{j\in \{OD, GD\}} \sum_{m\in {cls_{ob}}} Focal Loss(m^{i,j},m_{gt}^{i,j})  + \\
	&&  \sum_{i\in \Omega} \sum_{j\in {R}} \sum_{k\in {cls_{fc}, cls_{cc}}} \mathrm{BiFoc Loss}(k_j^{i}, k_{j,gt}^{i}) + \\
	&&  \sum_{i\in \Omega} \sum_{j\in {gd}} \sum_{k\in {\theta}} \mathrm{MSE}(k_j^{i},k_{j,gt}^{i})
\end{eqnarray*}
where $x, y, w, h, z$ are functions of $t^x, t^y, t^w, t^h, t^z$ respectively,
$\Omega$ is the grid cells where the object or grasping object are located. 
Since $cls_{fc}$ and $cls_{cc}$ are multi-classes, we used binary focal loss (BiFocLoss)~\cite{lin2017focal}. Focal loss gamma is set to 2 and we set  $\lambda_\mathrm{n} = 100$.
\section{EXPERIMENTAL EVALUATIONS}
We evaluated our proposed methods on the VMRD dataset~\cite{zhang2018rprg}, the Cornell dataset~\cite{Lenz:2013uz}, real robot grasping of single novel objects with a 4-axis robot arm and real multi-task robot grasping of multiple novel objects in various scenes with a 7-axis Baxter robot.
A RGB-D camera (Intel RealSense D435) was installed and used to have the field-of-view including robot and workspace from the top.
%In the robot experiment, Baxter was used for multi-task robot grasping and Dobot was used for real single object robot grasping. This is because Dobot's gripper is 27mm and Baxter's gripper is 75mm, so we can measure the model's sophistication with real robot grasping. In the real multi-task robot grasp experiment, we used Baxter to grasp various objects. 
\subsection{Implementation details}
Darknet-53 was also implemented used for the evaluations on the VMRD and for real multi-task robot grasping of multi objects. Darknet-19 was implemented and used for other evaluations. Either stochastic gradient descent (SGD) with momentum of 0.9 or Adam optimizer was used for training.
 %We used the backbone network with PyTorch for Darknet-19 and Darknet-50. We investigated SGD with momenum to 0.9 optimizer and Adam optimzer. 
Learning rate was 0.001 and mini batch size was set to 2.
For self-evaluation to optimize the model, total epoch was 50. Once the model is optimized, total epoch was set to 100 with reducing learning rate by half every 30 epochs.
% In self-evaluation, we set the epoch to 50 in VMRD Self-evalution to find the optimal model. After that, total epoch number was 100 with reudcing learning rate by half every 30 epochs on the optimal model.
Patch based training was performed with the sizes of $608\times608$ using data augmentation~\cite{zhang2018rprg}. 
%Note that backbone networks such as Alexnet, Darknet-19, Darknet-53, Resnet-50 and Resnet-100 require 61.1MB, 20.8MB, 41.6 MB, 25.5MB and 44.5MB memory and 3.1ms, 6.2ms, 13.7ms, 11.4ms and 20.0ms per image for image classification on a Titan X, respectively~\cite{darknet13}. 
All algorithms were tested on the platform with a single GPU (NVIDIA GTX1080Ti),a single CPU (Intel i7-7700K 4.20GHz) and 32GB memory.
 
\subsection{Evaluations on VMRD and Cornell datasets}

\begin{figure}[!t]
		\begin{center}	
        \begin{subfigure}[!b]{0.45\textwidth}
        		\centering
                \includegraphics[width=1.0\textwidth]{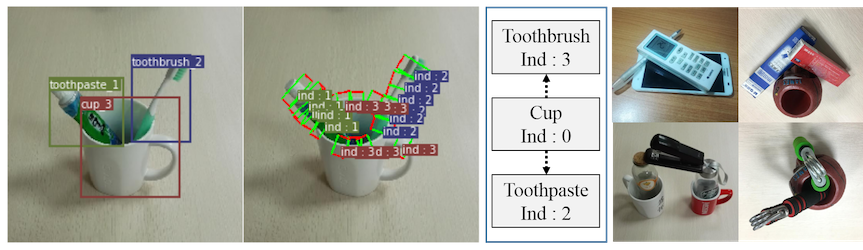}
                \caption{}
                \label{fig:vmrd}
        \end{subfigure}
        \begin{subfigure}[!b]{0.32\linewidth}
				\centering
                \includegraphics[width=1.0\linewidth]{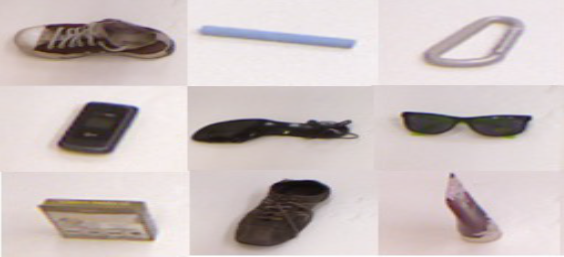}
                \caption{}
                \label{fig:cornelldb}
        \end{subfigure}
        \begin{subfigure}[!b]{0.32\linewidth}
        		\centering
    	\includegraphics[width=1.0\linewidth]{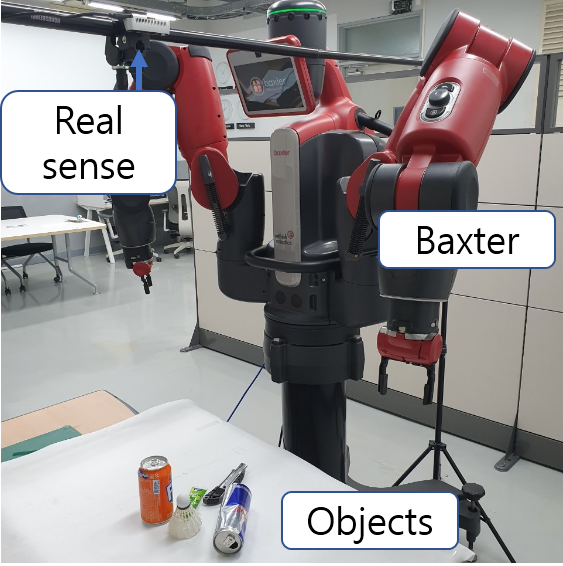}
                \caption{}
                \label{fig:baxter}
        \end{subfigure}
        \begin{subfigure}[!b]{0.32\linewidth}
        		\centering
    	\includegraphics[width=1.0\linewidth]{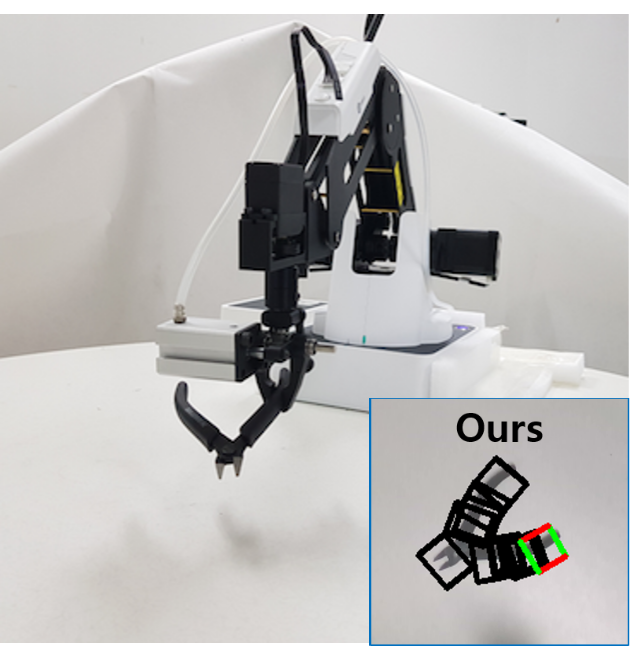}
                \caption{}
                \label{fig:dobot}
        \end{subfigure}
        \caption{(a) VMRD dataset. (b) images from Cornell dataset.  (c) our real multi-task evaluation environment (Baxter). (d) our robot grasping experiment with 4-axis robot.}
        	\label{fig:cornellresult1}
		\end{center}
			\vskip -0.15in
\end{figure}

We performed benchmarks using the Cornell dataset~\cite{Lenz:2013uz,Lenz:2015ih}
as illustrated in Fig.~\ref{fig:cornelldb}. 
This dataset consists of 855 images (RGB-D) of 240 different objects with GT labels 
of a few graspable / non-graspable rectangles. 
We cropped images with 360$\times$360, but did not resize it to 224$\times$224.
Five-fold cross validation (CV) was performed and average prediction accuracy was reported for image-wise and object-wise splits. %Image-wise split divides Cornell dataset into the train dataset and the test data with the ratio of 4:1 randomly without considering the same or different objects. Object-wise is a method of splitting train data and test data at a 4:1 ratio for each object.
When the difference between the output orientation $\theta$ and the GT orientation 
$\theta_{gt}$ is less than a certain threshold (e.g., 30$\degree$), then
IOU that is larger than a certain threshold ($e.g.$, 0.25, 0.3) will be 
considered as a successful grasp detection.
The same metric for accuracy has been used 
in previous works~\cite{Lenz:2015ih,Redmon:2015eq,Kumra:2017ko,Asif:2017bv,guo2017hybrid,Chu:ek,zhou2018fully,zhang2018roi}.

VMRD dataset was used to train our single multi-task network. VMRD consists of 4233 train data and 450 test data (RGB images) as illustrated in Fig.~\ref{fig:vmrd}. In this dataset, there are 2-5 objects stacked in each image and GT for OD with with class label \& relationship index, GD with class label and FC / CC labels. There are 31 object classes. 
% The measurement method is that i
If the IOU for predicted OD and GT OD is larger than 0.5 and the best grasping point for that object meets the above Cornell evaluation metric, it is considered as success (mAP with grasp or mAPg)~\cite{zhang2018rprg}.
 %우리는 Object detection with grasp detection에 대한 결과를 측정하기 위하여 이전 논문에서 사용한 방법을 사용하였습니다
 %VMRD에서 object detection label는 존재하지만, grasp detection label이 존재 하지 않을경우, 성공이라고 판단한 경우와 실패라고 판단한 경우에 대해서 결과를 report하였습니다.
%\subsection{Evaluation on Cornell Dataset}

\subsection{Evaluation of multi-tasks OD, GD, reasoning with Baxter}

 We evaluated our proposed methods using a Baxter with 7-axis arms (Rethink Robotics, Germany, see Fig.~\ref{fig:baxter}) for three different scenarios of 
 % The evaluations is consists of three ways. How does the robot can target grasping in
 cluttered scene, stacking scene and complex invisible stacking scene. 
 In cluttered scene, it was recorded as success if the robot grasped the target in a single try. In stacking and invisible scenes, it was recorded as success if the robot removed objects over the target and then grasp the target.
% if our method encounter a child on target then, put away that and grasping the target then it recorded as success. 
For the invisible scene with no target detected, the robot put away overlapping objects one by one until the target is found.
%the case there is no detection in our image for the target, then our method put away the overlapping one by one and grasping our target when it appear and predicted in network. 
Prediction was performed separately between all robot movements. All combinations of items, the target object and stacking orders are chosen randomly.
%in random orders and also target objects.

\subsection{Evaluation of GD with 4-axis robot arm}
We evaluated our proposed methods with a small 4-axis robot arm (Dobot Magician, Shenzhen
YueJiang Tech, China) for novel object grasping.
The following 8 novel objects (toothbrush, candy, earphone cap, cable, styrofoam bowl, L-wrench, nipper, pencil)
were used for grasping tasks.
If the robot gripper grasps an object and moves the object to another place, it is counted as success.

\section{RESULTS}

\subsection{Simulation results on VMRD dataset}

\begin{table}[!b]
\vskip -0.15in
\caption{Self-evaluation summary on VMRD.}
%We observe the best Map(\%) with grasp values on VMRD in 50 epochs.}
\label{tb2:VMRDresults1}
	\begin{tabular}{|c|c|c|c|c|}
		\hline
		\begin{tabular}[c]{@{}c@{}}Across scales\end{tabular} & Activation & Loss & Opt.    & \begin{tabular}[c]{@{}c@{}} mAPg (\%)\end{tabular} \\ \hline
		1, 2, 3                                                 & Sigmoid    & Cross Entropy  &Adam     & 56.5                                                          \\
		1, 2, 3                                                 & Softmax    & Cross Entropy   &Adam    & 63.1                                                          \\
		1, 2                                                    & Softmax    & Cross Entropy & Adam & 64.9                                                          \\
%		1                                                       & Softmax    & Cross Entropy  &Adam     & 62.7                                                          \\
%		2                                                       & Softmax    & Cross Entropy  &Adam     & 62.3                                                          \\
%		3                                                       & Softmax    & Cross Entropy   &Adam    & 57.4		\\
		1, 2                                                     & Softmax    & Focal Loss  & Adam & 67.1                                                          \\ 
		1, 2                                                     & Softmax    & Focal Loss  & SGD & 69.2                                                          \\ \hline
	\end{tabular}
	\label{tbl:selfVMRD}
%\vskip -0.1in
\end{table}

Table~\ref{tb2:VMRDresults1} summarizes our ablation study results on the VMRD multi-task robot grasp dataset. The method on the first row of Table~\ref{tb2:VMRDresults1} is an initial extension of YOLOv3 to multi-task robot grasping. Then, by changing activation function, scale, loss function and optimization algorithm, we were able to optimize our single DNN for multi-task OD, GD and reasoning from 56.5\% mAPg (mAP with grasp) to 69.2\% mAPg. Note that focal loss is often used for predicting unbalanced object classes~\cite{lin2017focal}.
%We conducted a total of four studies to use YOLO-v3 on the multi-task robot grasping. First of all, YOLO-v3 used sigmoid and binary cross-entropy loss for object class, but found that using the softmax and cross-entropy loss yielded improved performance. Second of all, we investigated the effect of our across scale approach for different scale prediction method of the grasp detection box. In the Across scale, what the numbers mean is the robot grasping detection output scale number. It was found to have high accuracy when using 1,2 on the across scale. Third of all, we improved the accuracy of the model by using Focal Loss~\cite{lin2017focal} to Cross-Entropy Loss to predict unbalanced object class. Finally, we change the Adam optimizer in YOLO-v3 for SGD with momentom optimizer to improve accuracy. 
%Using Softmax and Focal Loss, the accuracy is improved because the number of objects by class in 
The VMRD dataset seems unbalanced since % problem. T
there are 2061 notebooks and 93 chargers.
Focal Loss gives small weights to well-classified examples while gives large weights to some examples that are difficult to classify to focus on learning difficult examples. 
%Through this method, we improved the results because we made the unbalanced trained model more balanced. 

\begin{figure}[!t]
        \begin{center}
		\centering
        \includegraphics[width=0.48\textwidth]{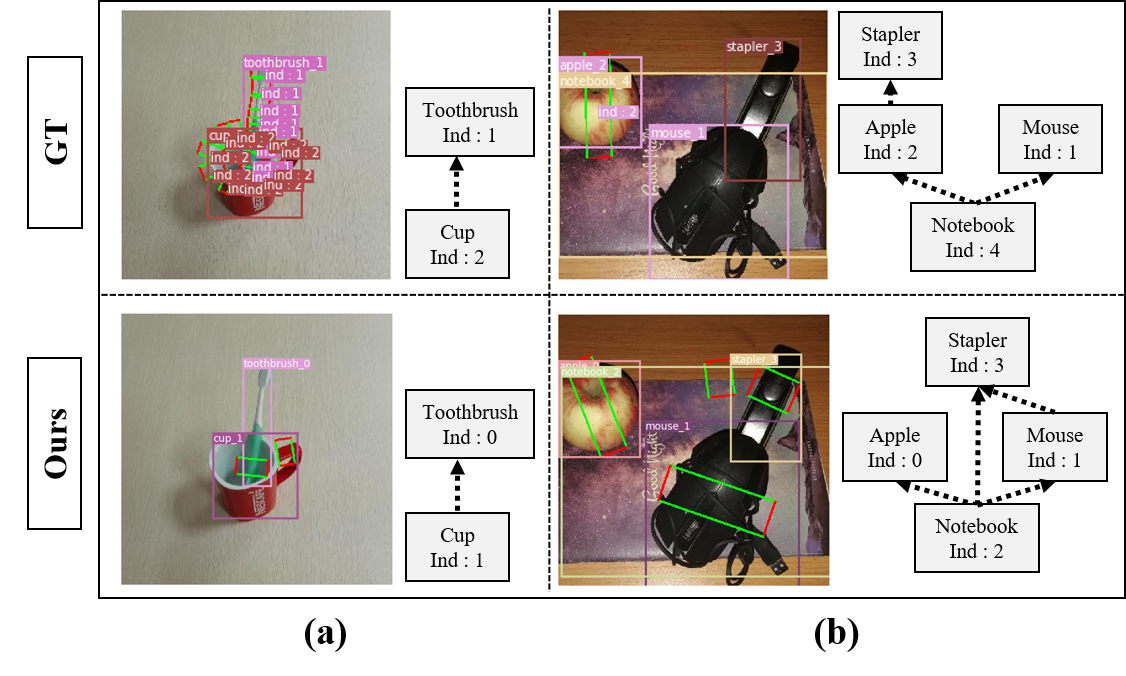}
        \caption{
        Multi-task detection results for VMRD. The 1$^\mathrm{st}$ row is GT and the 2$^\mathrm{nd}$ row is the results of our proposed methods. Note that our method
        yielded correct reasoning result for ``Stapler'' while GT incorrectly
        describes it.}
        %results about the robot grasping with object detection and reasoning.}
        \label{fig:vmrdResults}
        \end{center}
			\vskip -0.15in
\end{figure}

Fig.~\ref{fig:vmrdResults} illustrates qualitative results for generating multi-task robotic grasps. Fig.~\ref{fig:vmrdResults}(a) shows a two-level stacking case and its OD, GD and reasoning results of our proposed method (bottom row) and GT (top row). 
Fig.~\ref{fig:vmrdResults}(b) shows another multi-stacking case of GT (top) and the output of our proposed method (bottom). Note that GT contains an error in reasoning (Stapler is not on the Apple) while our method corrected for it through training on many examples.
%is ground truth error case of the grasp detection label and reasoning. 

Table~\ref{tbl:VMRDbench} summarizes the results of the results of previous methods~\cite{zhang2018rprg,zhang2018roi} and our proposed method. Our proposed method yielded state-of-the-art performance of 74.3\% mAP with grasp (mAPg) at the fastest computation speed of 33.3 FPS for a high resolution input image (608$\times$608). %Thus our proposed method. Thus, our proposed method produced better results than Zhang~\cite{zhang2018rprg,zhang2018roi} works.

\begin{table}[!h]
%\vskip -0.15in
\caption{Performance summary on VMRD dataset. 
%The Zhang methods~\cite{zhang2018rprg,zhang2018roi} consist of Resnet-101, and our network consists of Darknet-53. 
}
\centering
	\begin{tabular}{|l|c|c|}
		\hline
		Method           & \begin{tabular}[c]{@{}l@{}} mAPg (\%)\end{tabular} & \begin{tabular}[c]{@{}l@{}}Speed (FPS)\end{tabular} \\ \hline
		Zhang~\cite{zhang2018roi} baseline, OD, GD & 54.5                                                          & 10.3                                                  \\
		Zhang~\cite{zhang2018roi}, OD, GD          & 68.2                                                          & 9.1                                                   \\
		Zhang~\cite{zhang2018rprg}, OD, GD, reasoning    & 70.5                                                          & 6.5                                                   \\ \hline
		Ours, OD, GD, reasoning                 & 74.6                                                          & 33.3                                                  \\ \hline
	\end{tabular}
	\label{tbl:VMRDbench}
			\vskip -0.15in
\end{table}

\subsection{Simulation results on Cornell dataset}

\begin{figure}[!t]
		\begin{center}	
        \begin{subfigure}[!b]{0.45\textwidth}
				\centering
                \includegraphics[width=1.0\textwidth]{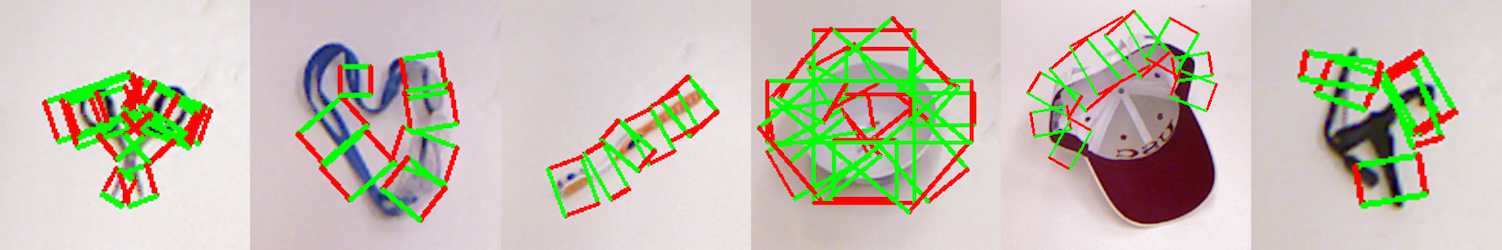}
                \caption{Ground truth}
                \label{fig:gull}
        \end{subfigure}
%          \begin{subfigure}[!b]{0.45\textwidth}
%        		\centering
%                \includegraphics[width=1.0\textwidth]{fig/base.png}
%                \caption{Redmon*}
%                \label{fig:gull2}
%        \end{subfigure}
        \begin{subfigure}[!b]{0.45\textwidth}
        		\centering
                \includegraphics[width=1.0\textwidth]{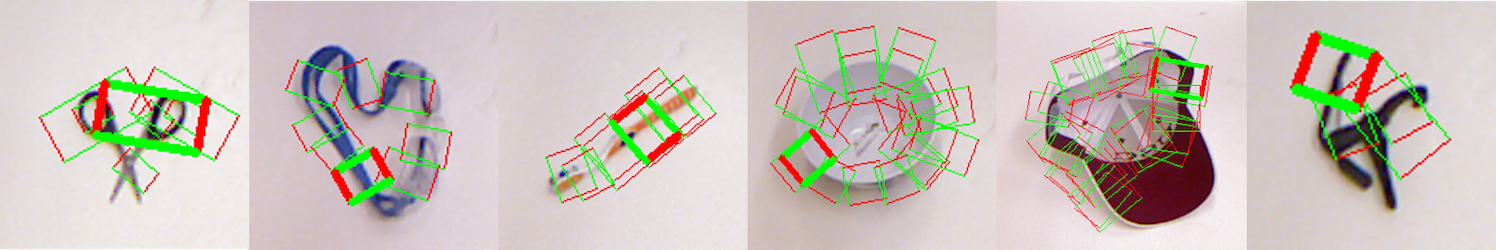}
                \caption{Ours without predictions across scales}
                \label{fig:gull2}
        \end{subfigure}
%        \begin{subfigure}[!b]{0.45\textwidth}
%        		\centering
%                \includegraphics[width=1.0\textwidth]{fig/ours_rgb.png}
%                \caption{Ours (RGB)}
%                \label{fig:tiger}
%        \end{subfigure}
        \begin{subfigure}[!b]{0.45\textwidth}
        		\centering
                \includegraphics[width=1.0\textwidth]{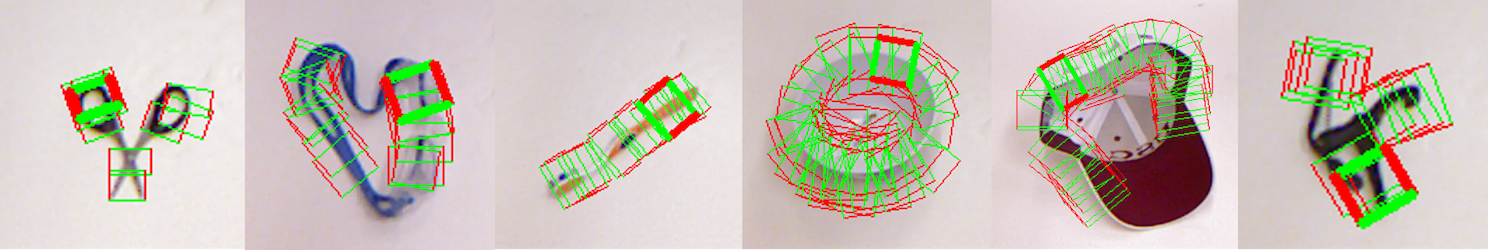}
                \caption{Ours with predictions across scales (proposed)}
                \label{fig:mouse}
        \end{subfigure}
        \caption{GD results on Cornell dataset using our methods without and with
        predictions across scales.}
%         Non across scale, and Ours(across scale) proposed methods with RGB.}
        	\label{fig:cornellresult1}
		\end{center}
			\vskip -0.15in
\end{figure}

Fig.~\ref{fig:cornellresult1} illustrates qualitative results for generating robotic grasps using our methods without and with predictions across scales. %) and using non across scale.
Both yielded fairly good grasp detection results, but there were often cases with fine details where predictions across scales improved the results such as the case with scissors as shown in Fig.\ref{fig:cornellresult1}.

Table~\ref{tbl:cornellresults1} summarizes the results of previous methods and our methods. Our proposed method with RGB-D yielded state-of-the-art performance of up to 98.6\% prediction accuracy for image-wise split and up to 97.2\% for object-wise split, respectively.
%, over reported accuracies of the work of Zhou~\cite{zhou2018fully} and others. 
Our proposed method with RGB also yielded comparable results to state-of-the-art methods. 
Note that our proposed method yielded these results with the smallest DNN and the fastest computation time of 16 ms per high resolution image (360$\times$360) that can be
potentially useful for real-time applications or stand-alone applications with limited memory and energy.
%Thus, our proposed method with RGB yielded better results than the work of Zhou~\cite{zhou2018fully} (except object-wise 25\%) with smaller network.
Using depth and predictions across scales % structures all yielded 
improved performance.

\begin{table}[!h]
\caption{Summary on Cornell data (25\% IOU).}
\label{tbl:cornellresults1}
\centering	\begin{tabular}{|l|c|c|c|c|}
		\hline
		 \multirow{2}{*}{Method}& \multirow{2}{*}{Input} & Image & Object    & Speed \\ 
		 
		         &         & (\%) & (\%) & (FPS)\\ \hline
		Lenz~\cite{Lenz:2015ih}, SAE         & RGB-D  & 73.9 & 75.6 & 0.08  \\
		Redmon~\cite{Redmon:2015eq}, Alexnet   & RG-D   & 88.0 & 87.1 & 13.2   \\
		Kumra~\cite{Kumra:2017ko}, Resnet-50  & RGB-D  & 89.2 & 88.9 & 16    \\
		Asif~\cite{Asif:2017bv}            & RGB-D  & 90.2 & 90.6 & 41    \\
		Guo~\cite{guo2017hybrid} \#a, ZFnet    & RGB-D  & 93.2 & 82.8 & -     \\
		Guo~\cite{guo2017hybrid} \#c, ZFnet    & RGB-D  & 86.4 & 89.1 & -     \\
		Chu~\cite{Chu:ek}, Resnet-50    & RG-D   & 96.0 & 96.1 & 8.3   \\
		Zhou~\cite{zhou2018fully}, Resnet-50   & RGB    & 97.7 & 94.9 & 9.9   \\
		Zhou~\cite{zhou2018fully}, Resnet-101  & RGB    & 97.7 & 96.6 & 8.5   \\
		Zhang~\cite{zhang2018roi}, Resnet-101 & RGB    & 93.6 & 93.5 & 25.2  \\ \hline
%		non across, Darknet-19  & RGB   & OSD & 96.4 & 94.0 & 140   \\
%		non across, Darknet-19  & RG-D  & OSD & 96.6  & 94.7 & 62.5  \\ 
		Ours, Darknet-19  & RGB    & 97.7 & 96.1 & 140   \\
		Ours, Darknet-19  & RG-D  & 98.6 & 97.2 & 62.5  \\ \hline
	\end{tabular}
			\vskip -0.15in
	\label{tbl:selfCornell}
\end{table}

%\begin{table}[]
%\vskip -0.15in
%\caption{Performance summary of reasoning on VMRD dataset.}
%\begin{tabular}{|c|c|l|l|}
%\hline
%Method    & Zhang~\cite{zhang2018visual} & Zhang~\cite{zhang2018rprg} & Ours \\ \hline
%Precision & 82.3  & 86.0  & 75.0 \\ \hline
%\end{tabular}
%\vskip -0.1in
%\end{table}

%
	
% Please add the following required packages to your document preamble:
% \usepackage{multirow}

\begin{figure}[!t]
			\vskip -0.1in
        \begin{center}
        \begin{subfigure}[!b]{0.311\linewidth}
				\centering
                \includegraphics[width=1.0\textwidth]{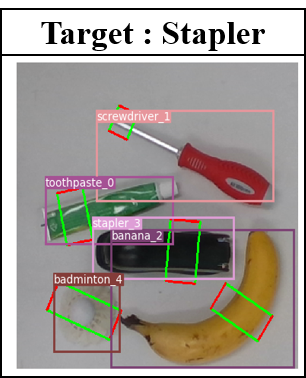}
                \caption{Cluttered scene}
                \label{fig:clutt}
        \end{subfigure}
        \begin{subfigure}[!b]{0.6\linewidth}
				\centering
                \includegraphics[width=1.0\textwidth]{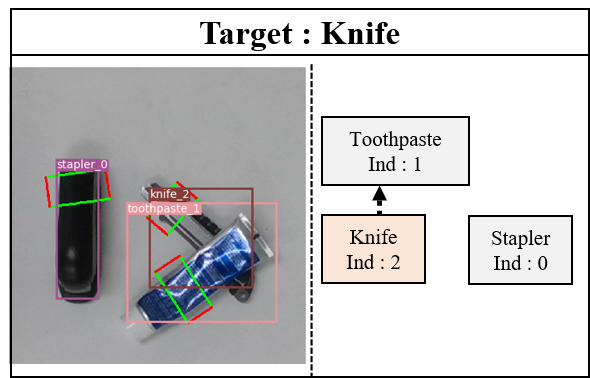}
                \caption{Stacking scene}
                \label{fig:stack}
        \end{subfigure}
        \begin{subfigure}[!b]{0.45\textwidth}
				\centering
                \includegraphics[width=1.0\textwidth]{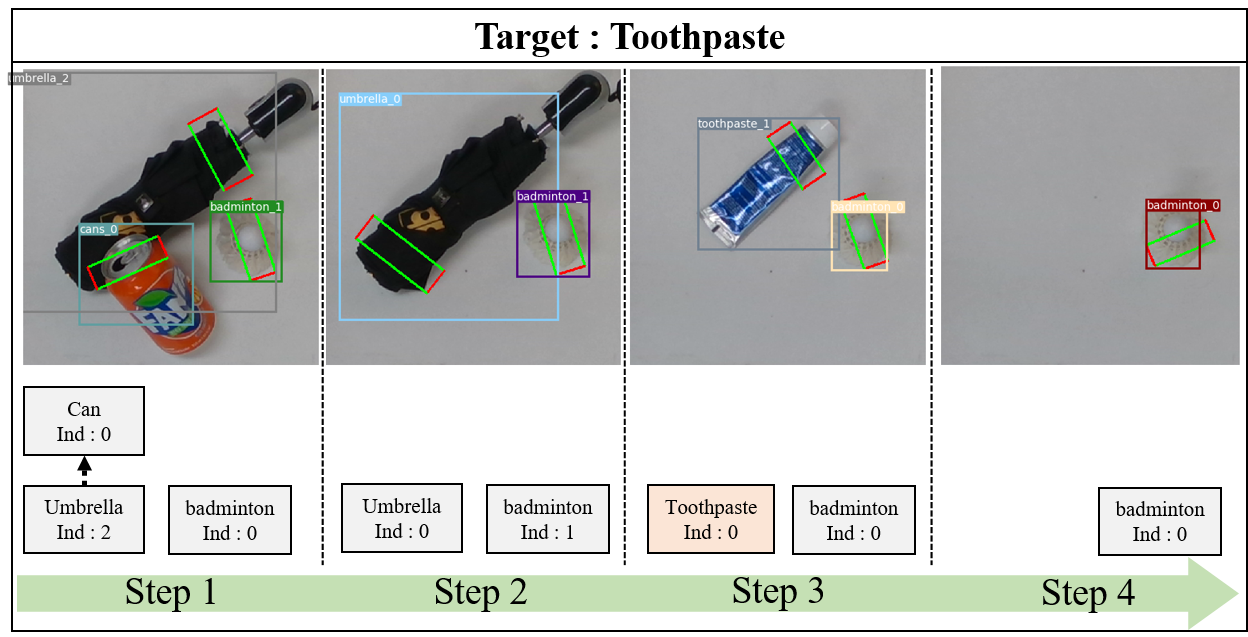}
                \caption{Invisible scene}
                \label{fig:invisible}
        \end{subfigure}
        \caption{Target grasp detection results in (a) cluttered scene, (b) stacking scene  and (c) challenging invisible scene.}
        \end{center}
			\vskip -0.15in
\end{figure}

%	In Figure\ref{fig:cornellresult1}, It seems that all methods yielded relatively good grasp  detection, but our proposed method (Ours) seems to predict the best robotic grasp0 information, close to the ground truth. Qualitatively,it was important to use No Class for accurate grasp-ability prediction.

\subsection{Results of multi-task OD, GD, reasoning with Baxter}

\begin{table}[!b]
			\vskip -0.15in
\caption{Performance summary of grasping tasks for cluttered (CS), stacking (SS) and invisible (IS) scenes.}
\begin{tabular}{|c|c|c|c|c|}
\hline
\#objects & 2            & 3            & 4             & 5             \\ \hline
CS      & -            & 86.7(13/15)  & 85.0\%(17/20) & 86.7\%(26/30) \\ \hline
SS      & 80.0\%(8/10) & 60.0\%(9/15) & 55.0\%(11/20) & -             \\ \hline
IS      & -            & 60.0\%(9/15) & 40.0\%(8/20)  & 28.0\%(7/25)              \\ \hline
\end{tabular}
\label{tbl:baxter}
\end{table}
  %The target grasping in various scenes in real can success well. 
  
  Fig.~\ref{fig:clutt} shows the OD, GD and reasoning results of our proposed methods for different scenarios such as (a) cluttered scene (CS), (b) stacking scene (SS) and (c) invisible scene (IS). For CS, the target ``Stapler'' was successfully located with proper grasp. For SS, the target ``Knife'' and its related object ``Toothpaste'' were well located with correct relationship reasoning. 
  For IS, the target was not detected due to occlusion, but as overlapped objects are removed based on the reasoning results (green arrow), the target was finally detected at the step 3.
  Table~\ref{tbl:baxter} shows the performance summary of the results of our proposed method with a Baxter robot. In CS, the accuracy was high, up to 86.7\% regardless of the increase in the number of objects. However, in SS, we observed that increasing the number of objects decreases grasp success rate possibly due to the difficulties of FC, CC predictions among them with severe occlusions. This phenomenon was also observed in challenging IS case.
  
%  Fig.~\ref{fig:stack} show that the results when the items are stacked. We target the knife at first step. After building the relationship among those objects, then firstly grasped the toothpaste and then re-try whole detection process and conduct grasping. However in the stacking scene, we found that increasing the number of objects had a significant effect on the accuracy, 
%thus making a lot of changes in the child and father class predictions. Fig.~\ref{fig:invisible} show that the results when target item is invisible. It also can be demonstrated well(60\%) but the same effect appear.

\subsection{Results of GD with 4-axis robot arm}

Fig.~\ref{fig:dobot} illustrates our robot grasp experiment with ``nipper''. 
Note that due to small gripper and small objects, grasp detection accuracy was important for
successful robot grasping.
Our proposed method yielded 95.3\% mean grasp success rate with 6.5\% standard deviation
for 8 novel, small objects with 8 repetitions per each object. 

\section{CONCLUSIONS}
We propose a single multi-task DNN that yields the information on GD, OD and reasoning among objects  with  a  simple  post-processing.  Our  proposed  methods yielded state-of-the-art performance with the accuracy of 98.6\% and 74.2\% and the computation speed of 33 and 62 FPS on VMRD and Cornell datasets, respectively.
Our  methods  also  yielded  95.3\%  grasp  success  rate  for  single novel object grasping with a 4-axis robot arm and 86.7\% grasp success  rate  in  cluttered  novel  objects  with  a  Baxter  robot.

%We proposed highly real-time, highly accurate robotic grasp detection methods that yielded state-of-the-art prediction accuracies  (up to 98.6\%)  with state-of-the-art computation times (16ms). We also proposed DNN based calibration method with 0.28mm validation error. Our proposed methods achieved 95.3\% success in real robotic grasping with challenging small gripper and small objects. We also demonstrate the importance and feasibility on grasping dynamic objects.

%\section*{APPENDIX}
%Appendixes should appear before the acknowledgment.

\section*{ACKNOWLEDGMENT}

This work was supported by the Technology Innovation Program or 
		Industrial Strategic Technology Development Program 
		(10077533, Development of robotic manipulation algorithm for grasping/assembling with 
		the machine learning using visual and tactile sensing information) funded 
		by the Ministry of Trade, Industry \& Energy (MOTIE, Korea).

\bibliographystyle{unsrt}
\bibliography{grasp}

\end{document}